%% file: main.tex
\documentclass{article}

\usepackage{arxiv}

\usepackage[utf8]{inputenc} 
\usepackage[T1]{fontenc}    
\usepackage{hyperref}       
\usepackage{url}            
\usepackage{booktabs}       
\usepackage{amsfonts}       
\usepackage{nicefrac}       
\usepackage{microtype}      
\usepackage{lipsum}		
\usepackage{graphicx}
\usepackage{natbib}
\usepackage{doi}

\usepackage{lscape}
\usepackage{amsmath}
\usepackage{amssymb}
\usepackage{xspace}
\usepackage{todonotes}
\usepackage{array}
\usepackage{hyperref}

\newcolumntype{R}{>{\raggedleft\arraybackslash}p{4em}}

\newcommand{\makesupplementtitle}[0]{\hsize\textwidth
    \linewidth\hsize \toptitlebar {\centering
        {\Large\bfseries Supplementary Material:\\\ Informed Priors for Knowledge Integration in Trajectory Prediction \par}}
    \bottomtitlebar}

\newcommand{\ade}[2][n]{%
    \ifx n#1\xspace{}minADE\(_{#2}\)\xspace \else
    \ifx b#1\xspace{}\textbf{minADE\(\mathbf{_{#2}}\)}\xspace 
    \fi\fi
    }
    
\newcommand{\fde}[2][n]{%
    \ifx n#1\xspace{}minFDE\(_{#2}\)\xspace  \else
    \ifx b#1\xspace\textbf{minFDE\(\mathbf{_{#2}}\)}\xspace 
    \fi\fi
    }

\newcommand{\hrate}[3][n]{%
    \ifx n#1\xspace{}HitRate\(_{#2,#3}\)\xspace \else
    \ifx b#1\xspace{}\textbf{HitRate\(\mathbf{_{#2,#3}}\)}\xspace
    \fi\fi
    }

\newcommand{\Tau}{\mathrm{T}}
\newcommand{\tpm}{$\pm$}

\title{Informed Priors for Knowledge Integration in Trajectory Prediction}

\author{\hspace{1mm}Christian Schlauch \\
	Humboldt-Universit\"at zu Berlin, \\ Chair of Statistics and Data Science\\
	and Continental AG\\
	Berlin, Germany\\ 
	\And
	\hspace{1mm}Nadja Klein \\
	Humboldt-Universit\"at zu Berlin\\ Chair of Statistics and Data Science \\
	Berlin, Germany
	\And 
	\hspace{1mm}Christian Wirth\\
	Continental AG\\
	Frankfurt am Main, Germany
}

\renewcommand{\shorttitle}{Informed Priors for Knowledge Integration in Trajectory Prediction}

\hypersetup{
pdftitle={Informed Priors for Knowledge Integration in Trajectory Prediction},
pdfsubject={q-bio.NC, q-bio.QM},
pdfauthor={Christian Schlauch, Nadja Klein, Christian Wirth},
pdfkeywords={Continual Learning, Informed Learning, Trajectory Prediction},
}

\begin{document}
\maketitle

\begin{abstract}
Informed machine learning methods allow the integration of prior knowledge into learning systems. This can increase accuracy and robustness or reduce data needs. However, existing methods often assume hard constraining knowledge, that does not require to trade-off prior knowledge with observations, but can be used to directly reduce the problem space. Other approaches use specific, architectural changes as representation of prior knowledge, limiting applicability. We propose an informed machine learning method, based on continual learning. This allows the integration of arbitrary, prior knowledge, potentially from multiple sources, and does not require specific architectures. Furthermore, our approach enables probabilistic and multi-modal predictions, that can improve predictive accuracy and robustness. We exemplify our approach by applying it to a state-of-the-art trajectory predictor for autonomous driving. This domain is especially dependent on informed learning approaches, as it is subject to an overwhelming large variety of possible environments and very rare events, while requiring robust and accurate predictions. We evaluate our model on a commonly used benchmark dataset, only using data already available in a conventional setup. We show that our method outperforms both non-informed and informed learning methods, that are often used in the literature. Furthermore, we are able to compete with a conventional baseline, even using half as many observation examples. 
\end{abstract}



\section{Introduction}

Deep learning approaches can achieve excellent prediction performance in many domains, but require large training data sets and can be brittle in open world applications \citep{lit:kiwissen}. However, training data may be difficult to obtain, e.g.~because relevant events are too rare or too costly to observe. Informed machine learning \citep{lit:informed} tries to alleviate this problem by integrating prior knowledge. This means, physical, world or expert knowledge is explicitly integrated into the learning system, and not just implicitly via observations. Doing so, allows to reduce the required training data, or increase accuracy, because less information needs to be derived from observations. Furthermore, the system can be more robust and safe, especially when encountering unseen events. \citep{lit:kiwissen}

Such an approach is of particular interest in the domain of autonomous driving, where applications have to reliably work under a wide variety of environments and safety-critical events, like near-collisions \citep{lit:kiwissen}. Furthermore, prior knowledge can be substantially cheaper in this domain, as it can be potentially defined with only map data \citep{lit:baseline}. An exemplary application is the trajectory prediction problem, that is especially difficult, since probabilities for multiple outcomes have to be predicted \citep{lit:tp_survey}.

We contribute to the area of informed machine learning and propose a probabilistic informed learning method based on continual learning \citep{lit:cl_survey}. The basic idea is to start with a learning task, that captures the prior knowledge. Then, in a second task when observed data is available, this prior knowledge can be turned into a weight posterior distribution, which can then be used as informed prior for the observation-based learning in future tasks. 
Such an approach has multiple advantages: (1) It is possible to integrate arbitrary prior knowledge, not just scientific knowledge, such as physical limits \citep{lit:kinematic,lit:conditional}. (2) We do not assume a specific representation of knowledge, only that it can be mapped to a prediction target. (3) The probabilistic formulation allows us to infer complete predictive distributions and achieve calibrated, multi-modal outputs beyond simple point predictions. Probabilistic formulations have been shown to lead to improved models, even if calibration and multi-modality only play a minor role \citep{lit:bayesian_ads}. (4) Furthermore, the approach allows us to potentially integrate multiple sources of prior information by learning multiple priors or by updating the prior sequentially.

As an exemplary illustration, we consider the trajectory prediction  and present a concrete implementation by adapting the state-of-the-art CoverNet trajectory predictor \citep{lit:covernet} with generalized variational continual learning \citep[GVCL;][]{lit:gvcl} and by using prior knowledge about the drivable area \citep{lit:baseline}. 

In multiple sets of experiments, we critically analyze our approach and show that it improves on existing non-informed baselines and other informed learning approaches like multi-task or transfer learning \citep{lit:baseline}.

In summary, our contributions are:
\begin{itemize}
    \item A novel approach for integrating prior knowledge as soft constraint into deep learning models, that can also deal with multi-modal and probabilistic prediction problems, distinct datasets and does not require architectural changes;
    \item an implementation that substantially improves the CoverNet trajectory predictor, only using data already available in the conventional setup; 
    \item a demonstration of the advantages of our informed machine learning method in the domain of trajectory prediction for autonomous driving;
    \item an empirical assessment of the different facets of our probabilistic approach.
\end{itemize}

The paper is structured as follows: In Sec.~\ref{sec:method} we introduce our informed machine learning approach before we exemplify it on the CoverNet architecture in Sec.~\ref{sec:covernet_gvcl}. Related work is covered in Sec.~\ref{sec:related_work}. In Sec.~\ref{sec:experiments} we benchmark our method against various baseline models and close the paper with a conclusion and discussion in Sec.~\ref{sec:conclusion}.

\section{Continual Learning with Prior Knowledge Integration}
\label{sec:method}

We introduce a regularization-based continual learning approach to explicitly integrate prior knowledge in the learning algorithm, aiming to increase the accuracy, robustness and safety of deep learning models, while making minimal assumptions on the type of prior knowledge or model architecture. In particular, we consider world and expert knowledge as defined in \citet{lit:informed} as our prior knowledge sources. In contrast to scientific knowledge, this knowledge can be violated and it is required to trade-off in form of \textit{soft constraints}.

\subsection{The Informed Learning Problem}

 Our focus is on  supervised learning problems, where we denote $\mathcal{D}=\{\mathcal{X}\times\mathcal{Y}\}$ as the domain over the input space $\mathcal{X}$ and the output space $\mathcal{Y}$. From this, we draw a labelled dataset $D=\{(x_i,y_i)\}^N_{i=1}$ consisting of a finite number of $N$ conditionally independent  realizations $y_i|x_i$, where $ x_i\in \mathcal{X}$ and $y_i \in \mathcal{Y}$. Let $p(y|x)$ denote the true conditional distribution of $Y$ given $X=x$ and let $\mathcal{H}$ denote a hypothesis space consisting of hypothesis (models) $h_\theta(\cdot)$ that are defined through some model parameters $\theta$. Given $D$ and $\mathcal{H}$, a learning algorithm $\mathcal{L}: D \to \mathcal{H}$ is applied to learn a final hypothesis $h_\theta(x) \in \mathcal{H}$, either in form of a predictive function $y = h_\theta(x)$ or in a probabilistic approach in form of a predictive distribution $h_\theta(x) \sim p_\theta(y|x)$. The goal is now to integrate the prior knowledge source $\mathcal{B}$ in possibly any of the elements $D$, $\mathcal{L}$ and $\mathcal{H}$ or in revision of $h_\theta(x)$ itself, thereby placing an inductive bias into the learning problem that increases accuracy, robustness, parsimony or compliance \citep{lit:kiwissen, lit:informed}. Following \citet{lit:informed}, we call the explicit integration of a prior knowledge source $\mathcal{B}$ 
 an \emph{informed learning problem}.

\subsection{Informed Priors with Continual Learning}

\begin{figure*}[t]
\includegraphics[width=\linewidth]{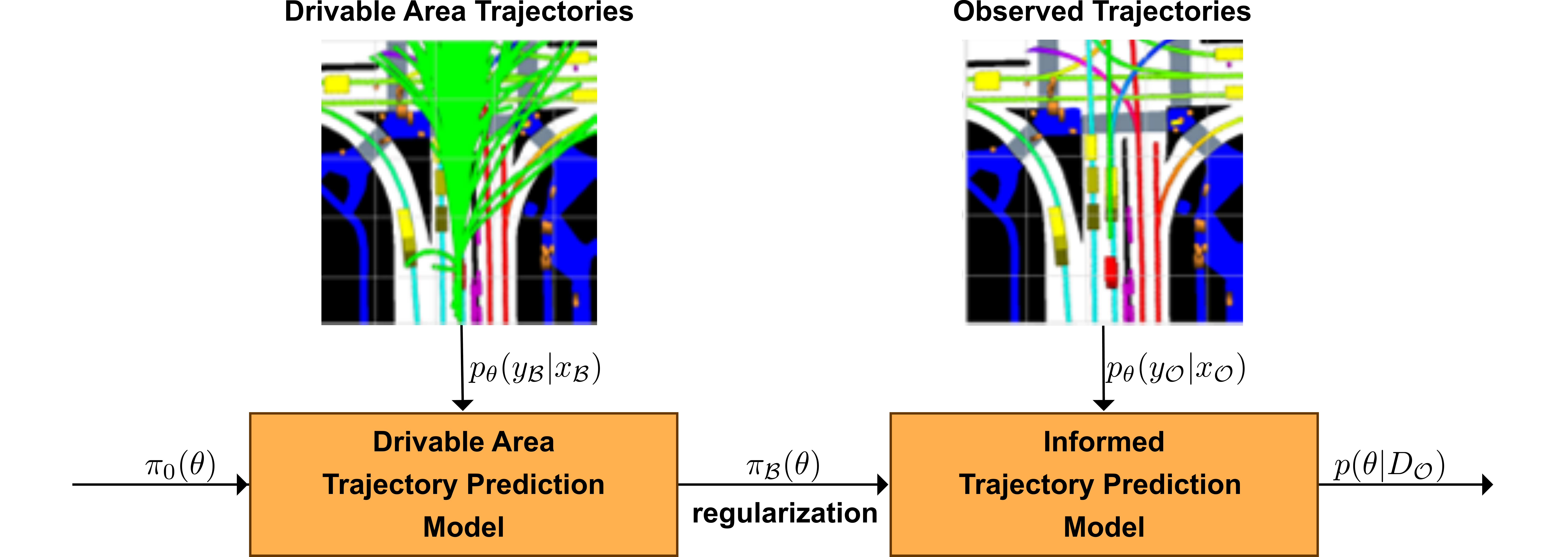}
\caption{Example for the integration of prior knowledge with continual learning: Using the drivable area as informed prior for trajectory prediction.}
\label{fig:informed_cl}
\end{figure*}

In a continual learning scenario we generally assume to sequentially encounter $t=1,\ldots,T$ disjoint, labelled training datasets (tasks) $D_t=\{(x^{(t)}_i, y^{(t)}_i)\}^{N_t}_{i=1}$ with $N_t$ independent samples each. These datasets  may potentially have been drawn from $T$ different domains $\mathcal{D}_t$ with different input spaces $\mathcal{X}_t$, output spaces $\mathcal{Y}_t$ under distinct conditional probability distributions $p_t(y|x)$ \citep{lit:cl_review, lit:cl_survey}. However, we assume that each dataset represents a related task and the general goal is to learn a single model that is able to predict well when seeing data from any of them \citep{lit:cl_survey}. In particular, we want to consider a probabilistic deep learning model with parameters $\theta$ that returns a predictive distribution $p_\theta(y| x)$ over an output $y$ given an input $x$. Placing a prior distribution $\pi(\theta)$ on $\theta$ allows to recursively apply Bayes' rule over all tasks given each likelihood $p_{\theta}(\cdot|x)$ to recover the posterior probability distributions $p(\theta|D_{1:T})$ with
\begin{equation}\label{eq:Bayes}
\begin{split}
    p(\theta| D_{1:T}) 
    \propto \pi_0(\theta) \prod^T_{t=1}p_{\theta}(y_t|x_t).
    \end{split}
\end{equation}
and $\pi_0(\theta)$ as the initial, uniformed prior. As uninformed prior, we use a zero-mean, unit-variance normal distribution.

Applying this approach to an informed learning problem, we assume that a given prior knowledge source $\mathcal{B}$ can be represented in form of a dataset $D_\mathcal{B}$. We can then define a task $t=1$ to learn the corresponding posterior distribution $\pi_{\mathcal{B}}(\theta)$ which constitutes an implicit probabilistic representation of $\mathcal{B}$. Now, given an observational dataset $D_\mathcal{O}$ in a second task $t=2$, we can use $\pi_{\mathcal{B}}(\theta)$ as an informed prior and regularize the posterior distribution $p(\theta| D_\mathcal{O})$ as $p_{\theta}(y_\mathcal{O}|x_\mathcal{O})\pi_{\mathcal{B}}(\theta)$. We present an exemplary visualization of this approach in Fig.~\ref{fig:informed_cl} for the trajectory prediction in autonomous driving considered later. 

This regularization-based approach implicitly trades-off the effect of $\pi_{\mathcal{B}}(\theta)$ with the available data in $D_\mathcal{O}$, making it suitable to consider world or expert knowledge as soft constraint. 
The less observational data is available, e.g.~for rare events in the input space $\mathcal{X}$, the more relevant the informed prior becomes. In theory, we may consider even multiple knowledge sources $\mathcal{B}_t$ sequentially, constructing a knowledge pipeline. The separation of concerns in such a sequential approach allows to reuse posteriors $\pi_{D_{1:t-1}}(\theta)$ as informed priors of the pipeline, to expand the latter when new data becomes available or to reorder the tasks to form a curriculum \citep{lit:cl_wholistic}.

In general, this approach does not require a specific model representation and can be applied to existing state-of-the-art models using a shared output representation for all tasks. We also note that the assumed dataset $D_\mathcal{B}$ can be derived from different knowledge representations, e.g.~by generating artificial examples in a simulation or by making semantic annotations in an observational dataset. In theory, due to the implicit trade-off in the approach, the number of samples required in $D_\mathcal{B}$ does not necessarily have to stand in balanced relation to the number of samples in $D_\mathcal{O}$.

\subsection{Societal Impact}

In many domains, we are confronted with implicit, societal biases in the training data; e.g.~we may observe drastic differences in driver's behavior between countries. In theory, our approach can be used to adapt a model by integrating available prior knowledge about social norms or regulations that determine such behavior. 
Furthermore, our approach may be used to validate a given model by testing its performance in a prior knowledge task, e.g. to check whether it complies to certain regulations or exhibits certain social norms or biases. We presume that such validation may become necessary for the certification of e.g.~autonomous driving systems in the conceivable future. Therefore,  our approach may have a positive impact. However, like any automatization efforts, advances in autonomous driving may lead to a reduction of available jobs. 

\section{Generalized Variational Continual Learning-CoverNet for Trajectory Prediction}
\label{sec:covernet_gvcl}

\begin{figure*}[t]
\includegraphics[width=\linewidth]{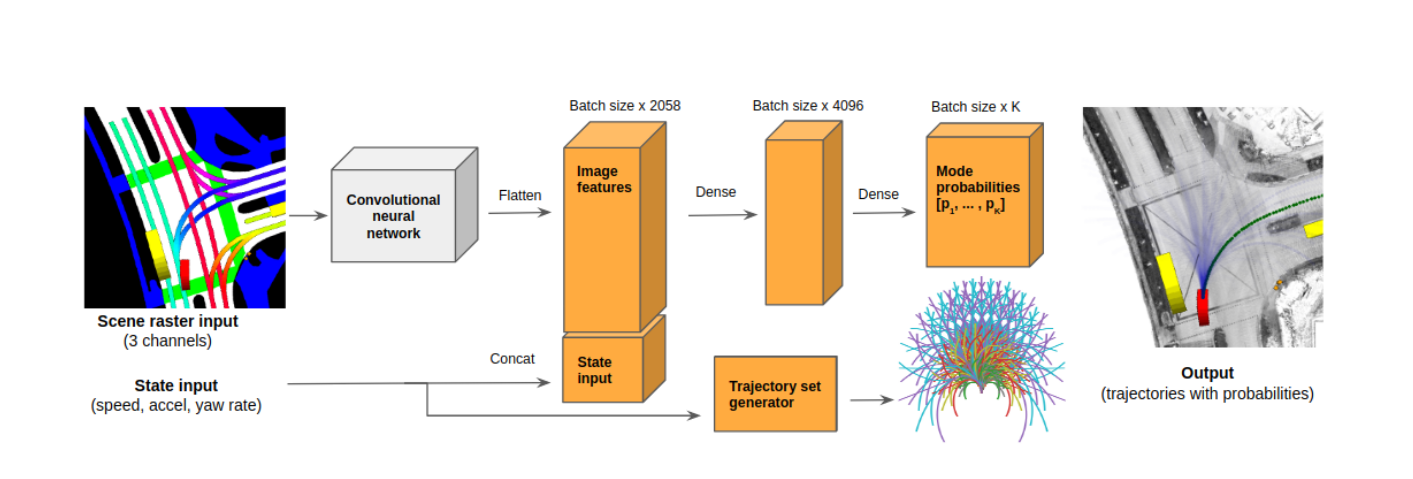}
\caption{CoverNet architecture from \cite{lit:covernet}.}
\label{fig:covernet}
\end{figure*}

We exemplify our continual learning approach by applying it to CoverNet \citep{lit:covernet}, a state-of-the-art trajectory prediction model \citep{lit:tp_survey_deep, lit:tp_survey}. We use drivable area map data as a prior knowledge source, which is also applied by \citet{lit:baseline} and allows comparisons to existing baselines. The trajectory prediction in autonomous driving is a safety-critical application \citep{lit:kiwissen, lit:tp_survey_deep} that renders the use of probabilistic multi-modal predictions useful to obtain calibrated probabilities along with accurate uncertainty measures for safety-critical events. 

\subsection{The Trajectory Prediction Problem}

 Consider a scene on the road, where a self-driving system is assumed to observe  states $s_i$ in the state space $\mathcal{S}$ of all involved agents $a_i\in\mathcal{A}$. 
 Let $s_i^{(\tau)} \in \mathcal{S}$ denote the state of agent $a_i \in \mathcal{A}$ at time $\tau$ and let $s_i^{(\tau-\Tau_o:\tau)}=\lbrace s_i^{(\tau-\Tau_o)}, s_i^{(\tau-\Tau_o+1)}, \ldots, s_i^{(\tau)}\rbrace$ be its trajectory over an observation period $\Tau_o$. Additionally, we assume access to map data $M \in \mathcal{M}$ and some available prior knowledge $B \in \mathcal{B}$ that can be translated into a prior $\pi_{\mathcal{B}}(\theta)$. The map data itself may include semantic information, e.g.~road markings and traffic signals, that can be used as further prior knowledge source. The scene context is then denoted as $\mathcal{C}=\{\{s_i^{((\tau-\Tau_o):\tau)}\}^{|\mathcal{A}|}_{i=1}, M, B\}$. Given $\mathcal{C}$, the goal is to predict the distribution of future trajectories $p(s_i^{(\tau:(\tau+\Tau_h))}|\mathcal{C})$
 over the prediction horizon $\Tau_h$.

\subsection{CoverNet informed by Driveable Area Knowledge}
\label{sec:covernet}
CoverNet as introduced by \cite{lit:covernet} solves the trajectory prediction problem, using a birds-eye-view RGB rendering that combines map data and states of all involved agents over a three second observation period. The output is represented by sparse trajectory sets $\mathcal{K}(\epsilon)$ that approximate $p(s_i^{(\tau:(\tau+\Tau_h))}|\mathcal{C})$ by a predefined set of possible trajectories. The sparsity is controlled by a coverage bound $\epsilon$ in meter, which determines the maximum distance between two trajectories in the set and thus, the total number of elements $|\mathcal{K}(\epsilon)|$ in the set. \cite{lit:covernet} computes $\mathcal{K}(\epsilon)$ by clustering all observed trajectories in a given dataset, with a cluster distance of $\epsilon$. This output representation allows to treat each trajectory in the set as a sample of the predictive distribution, while only computing its conditional probability. This allows the authors to reduce the task to a classification problem, with the goal to approximate the probability of each class (or trajectory distribution sample). The architecture is visualized in Fig.~\ref{fig:covernet}.

We implement our approach in Sec.~\ref{sec:method} by applying generalized variation continual learning \citep[GVCL;][]{lit:gvcl} to CoverNet. GVCL is a likelihood-tempered version of variational continual learning \citep[VCL;][]{lit:vcl} based on variational inference \citep[VI;][]{lit:bayes_temper}. The likelihood-tempering has been shown to help in VI \citep[][]{lit:bayes_temper}, in some settings. GVCL solves the recursion described in Eq.~\eqref{eq:Bayes} by minimizing the likelihood-tempered log-evidence lower bound (so-called $\beta$-ELBO, $\beta \in [0,1]$) at every task $t=1,\ldots,T$. This yields a tractable approximate solution $q_t(\theta) \approx p(\theta| D_{1:t})$ using the approximate posterior $q_{t-1}(\theta) \approx p(\theta |D_{1:t-1})$ of the previous task $t-1$ as a prior distribution weighted by $\beta$ for all $t$. More precisely, the $\beta$-ELBO at time $t$ is defined by
\begin{align}\label{eq:gvcl}
    \beta\text{-ELBO} = \mathbb{E}_{\theta \sim q_t(\theta)}\left[\log{p_\theta(y_{t} | x_{t})}\right]- \beta D_{\text{KL}}(q_t(\theta) \| q_{t-1}(\theta)),
\end{align}
where $D_{\text{KL}}(q_1||q_2)$ is the Kullback-Leibler (KL) divergence between two probability distributions $q_1,q_1$. We assume a fixed form  $q_t(\theta)$ that belongs to the family $\mathcal{Q}$ of Gaussian distributions with diagonal covariance matrices. In our GVCL-CoverNet architecture, $\theta$ collects as usually all weights and biases, and the variational parameter vector $\nu=(\mu,\sigma)^\top$ consists of the means and standard deviations of the independent Gaussian variational distributions of each model weight.

We inform the trajectory prediction by placing a soft constraint on on-road trajectories. 
Thus, in the first task we assume that drivable area map data is available in the scene context $\mathcal{C}$. Given this map data, we define a multi-label classification where the positive labels are given by a binary vector $r_i^{(\tau)} \in \mathbb{R}^{|\mathcal{K}|}$ for an agent $a_i$ at time $\tau$, defined such that its value is 1 at entry $k_{i,j}^{(\tau)}$ if the $j$th tracetory $k_{i,j}^{(\tau)}$ in the set $\mathcal{K}$ is entirely contained in the drivable area and 0 otherwise. The model is then trained using a binary cross-entropy loss. The resulting posterior is used in the second task to regularize the model according to Eq.~\eqref{eq:gvcl}. Given the observational dataset $D_\mathcal{O}$, this task is defined as a multi-class classification with positive samples determined by the element in the trajectory set closest to the actual ground truth. The model is finally trained using a sparse categorical cross-entropy loss where its logit transformations are normalized by the softmax function, inline with the original CoverNet algorithm.

Note, even though we cannot enforce the prior knowledge as hard constraint, we may artificially sharpen the informed prior in the $D_{\text{KL}}$-term as described by \citet{lit:gvcl}. This allows to strengthen the effect of the informed prior, making GVCL a particularly flexible method choice. Furthermore, GVCL can be applied to wide variety of model architectures in general \citep{lit:vcl, lit:gvcl}, with the exception of recurrent architectures for which variational reformulations are typically more involved \citep{lit:recurrent}.

\section{Related Work}\label{sec:related_work}

We briefly embed our contribution into the literature by reviewing related work next. A general survey on informed machine learning can be found in \citet{lit:informed} providing a taxonomy based on the type of knowledge source, representation and integration. Prior knowledge integration in the domain of autonomous driving is reviewed by e.g.~\citet{lit:kiwissen}. 

In the trajectory prediction for autonomous driving, there has been a particular focus on integrating scientific knowledge about physical limitations using plausible dynamical models. For instance, \citet{lit:covernet} propose dynamically constructed trajectory sets as output representation, while \citet{lit:kinematic} and \citet{lit:conditional} define specific layer types to directly encode the dynamical model. In contrast, \citet{lit:mpc} present a post-processing approach based on a model predictive controller. While these methods enforce the physical limitations as hard constraints, we enable the use of less formalized world and expert knowledge as soft constraint. Earlier work in this direction has been done by \citet{lit:offroad_regression} using a multi-task approach, while \citet{lit:baseline} introduce a loss based method combined with a transfer learning approach where the model is pre-trained using prior knowledge. Similar approaches have also been shown to be beneficial in other domains \citep{lit:informed_pretrain}. However, transfer learning is typically limited to an unidirectional knowledge transfer between two tasks, while multi-task learning and loss approaches require a single dataset with simultaneously available labels. A continual learning approach can be applied without these limitations. To our knowledge, a continual learning approach has not yet been used to realize an informed machine learning algorithm.

Continual learning methods are typically categorized as regularization-based, rehearsal-based or as architectural \citep{lit:cl_review, lit:cl_survey}. Architectural methods typically require a specific model representation, while rehearsal-based methods require a so-called replay buffer. Regularization-based methods do not have these requirements and can be easily formulated in a probabilistic way, which is itself argued to feature robustness and safety \citep{lit:bayesian_ads}. We use the VI-based GVCL method, which generalizes variational continual learning \citep[VCL;][]{lit:vcl}  and online elastic weight consolidation \citep[Online EWC;][]{lit:ewc, lit:online_ewc}.


\section{Experiments}
\label{sec:experiments}
We empirically investigate the performance of the proposed GVCL-CoverNet model using the NuScenes dataset of \citet{lit:nuscenes}. This dataset contains 1000 scenes, each being 20 seconds, sampled at a frequency $F=2\text{ Hz}$. In the following, we introduce two non-informed and two informed baseline methods as competitors. Thereafter, we present the experimental design, the implementation details and lastly the results.

\subsection{Baseline Models}
\label{sec:baselines}

We consider two non-informed baseline models:
\begin{itemize}
    \item \textbf{Base-CoverNet}, as introduced by \citet{lit:covernet} and reviewed in Sec.~\ref{sec:covernet_gvcl}. All other models used in our experiments are adaptations of this model.
    \item \textbf{VI-CoverNet} is a variational adaptation  of the Base-CoverNet model. Like our GVCL-CoverNet model described in Sec.~\ref{sec:covernet_gvcl}, the number of parameters is doubled, in comparison to \textbf{Base-CoverNet}, and the likelihood-temperature is set by a hyperparameter $\beta \in [0,1]$. The consideration of this baseline allows us to distinguish the effects of the continual learning approach and the variational formulation alone.
\end{itemize}
We also consider two informed baseline models that enforce on-road trajectory predictions as soft constraint:
\begin{itemize}
    \item \textbf{Loss-CoverNet} uses an auxiliary off-road loss as introduced by \cite{lit:baseline}. The model jointly trains on the prior knowledge and the observations. The trade-off between both losses is controlled by a hyperparameter $\lambda_\text{\scriptsize{multi}} > 0$. 
    \item \textbf{Transfer-CoverNet} uses a model pre-trained using prior knowledge, similar to \citet{lit:baseline}.
\end{itemize}
Since our GVCL-CoverNet model has mean and scale as variational parameters, we also introduce \textbf{GVCl-Det-CoverNet} to allow for a fair comparison to the non-variational baselines at test time. The GVCL-Det-CoverNet is still trained according to the GVCL method introduced in Sec.~\ref{sec:covernet_gvcl} but makes non-probabilistic predictions at test time using the mean $\mu$ of $\nu$ only. 

\subsection{Experimental Design}
\label{sec:experiment_design}
In total, we run three different sets of experiments with 10\%, 50\% and 100\% of the available observations in the training set. However, we use all available scenes for determining the drivable area prior.
This relates to real world driving applications, where map data is substantially cheaper to obtain than observations and where thus  prior knowledge is cheaper to assess than observations. 

We use the data split described by \citet{lit:covernet} with 32186 observations in the full training set, 8560 observations in the validation set and 9041 observations in the test set. The 10\% and 50\% subsets are randomly sub-sampled from the training data, while using the full validation and test set.


In our first set of experiments, we use the fixed trajectory sets $\mathcal{K}(\epsilon)$ from the NuScenes prediction challenge \citep{lit:nuscenes} with the coverage bound $\epsilon=4$, corresponding to $|\mathcal{K}(4)|=415$ modes. In a second part, we investigate the influence of the coverage bound and additionally consider the bounds $\epsilon \in \{2, 8\}$, corresponding to $|\mathcal{K}(2)|=2206$ and $|\mathcal{K}(8)|=64$ modes, respectively.

To evaluate each experiment, we consider the following metrics that have been commonly applied in the literature \citep{lit:tp_survey, lit:tp_survey_deep}. The average displacement error \ade[n]{1} and final displacement error \fde[n]{1} evaluate the quality of the most likely trajectory, whereas \ade[n]{5} considers the five most likely trajectories. Please note, that these metrics are limited by the resolution of the trajectory set $\mathcal{K}$. Therefore, we also report the rank (RNK), determining the position of the best trajectory, when ranking the predictions by probability. Top-1 accuracy (ACC) determines how often the best trajectory from the trajectory set is predicted with the highest probability. The negative log likelihood (NLL) and expected calibration error (ECE) allow to evaluate the calibration of the probabilistic multi-class prediction. The drivable area compliance (DAC) metric determines in how far the predictions conform to the prior knowledge. The \ade[n]{5} and the RNK metric indirectly depend on well calibrated, probabilistic predictions, as they evaluate the quality of the probability-based ordering.

\subsection{Implementation Details}

All presented models have the same architecture, as described by \citet{lit:covernet}. However, we do not use ImageNet pre-training, as we do not observe any improvements on the baseline model and there are no ImageNet weights readily available for the variational models. Instead, we initialize the deterministic weights $\theta$ and variational means as proposed by \citet{lit:henorm}. The variational standard deviations $\sigma$ are drawn from the Gaussian distribution $\mathcal{N}(0.005, 0.1)$ truncated at $(0.005,0.205)$. The deterministic models are trained with stochastic gradient descent \citep[SGD;][]{lit:sgd} over 20 epochs. The variational models are trained by minimzing the respective ELBO with 
the reparameterized gradient method \citep[][]{lit:repara} over 2000 epochs and linearly decaying learning rate. We slightly temper the posterior of the variational models by setting $\beta$ to 1 divided by batch size, following \citet{lit:bayes_temper}. At test time, the variational models average over 7 samples.

We individually tune the learning rates and batch sizes for each model as well as the $\lambda_\text{\scriptsize{multi}}$ for the Loss-CoverNet baseline model using 100\% of the training data and coverage bounds $\epsilon \in \{2,8\}$. We adopt the tuning results for $\epsilon=8$ for the experiments with $\epsilon=4$. The values are selected to minimize the validation NLL, using the heteroscedastic and evolutionary Bayesian optimizer \citep[HEBO][]{lit:hebo}. The chosen values are summarized in Tab.~\ref{tab:hypparams}.


The models have been implemented in Tensorflow \citep{lit:tensorflow}, published on Github\footnote{\url{Anonymized}}. The experiments have been executed on a high performance computation cluster with 4 NVIDIA GeForce RTX 3090-24GB GPUs and NVIDIA V100-32GB GPUs with parallelization for 4 GPUs.

\begin{table}[h]
\footnotesize
\caption{Selected values for tuned hyperparameters.} \label{tab:hypparams}
\begin{center}
\begin{tabular}{lrrr}
\textbf{Method} & \textbf{Batch Size} & \textbf{Learning Rate} &\textbf{$\lambda_{\scriptsize{\text{multi}}}$}\\
\hline \\
$\epsilon$=2  & &   & \\  
\hspace*{3mm} Base       &  5  & 0.00083    &  \\
\hspace*{3mm} Transfer &  5   &  0.00083   & \\
\hspace*{3mm} Loss        & 5 & 0.00083  &  0.01 \\
\hspace*{3mm} VI        &   12 & 0.003  &    \\
\hspace*{3mm} GVCL  &   12 & 0.003  &  \\ \\
$\epsilon$=4  & &   & \\      
\hspace*{3mm} Base       &  16  &     0.0008  & \\
\hspace*{3mm} Transfer &   16   &    0.0002 &     \\
\hspace*{3mm} Loss        &  16  &  0.0001& 0.01\\
\hspace*{3mm} VI        &   12 & 0.003  &    \\
\hspace*{3mm} GVCL  &   12 & 0.003  &  \\ \\
$\epsilon$=8  & &   & \\                
\hspace*{3mm} Base     &   16  & 0.0008& \\
\hspace*{3mm} Transfer & 16  & 0.0002&\\
\hspace*{3mm} Loss        &  16  & 0.0001 & 0.01 \\
\hspace*{3mm} VI        &   12 & 0.003  &    \\
\hspace*{3mm} GVCL  &   12 & 0.003  &
\end{tabular}
\end{center}
\end{table}

\begin{table*}[t]
\footnotesize
\caption{Results for constrained observational training datasets with coverage bound $\epsilon=4$ (\textbf{bold} as unique best result and \textit{italic} as unique best deterministic per set)} \label{tab:results_partial}
\begin{center}
\begin{tabular}{l*{8}{R}}
\textbf{Experiment} & \textbf{ACC} & \textbf{NLL} & \textbf{ECE} & \textbf{RNK} & \ade[b]{1} & \ade[b]{5}& \fde[b]{1} & \textbf{DAC} \\
\hline \\
100\% Data & & & & & & & & \\
\hspace*{3mm} Base  &  0.15  &  3.40        &  \textit{0.05}      &  15.07  &  4.77     &  2.26      & 10,61    &  0.88\\
\hspace*{3mm} Transfer &  0.15   &  \textit{3.24} & 0.06 & \textit{11.56}    &   4.70  &  \textit{2.12}  & 10.15   & \textbf{\textit{0.96}} \\
\hspace*{3mm} Loss        &  0.13  &  3.58        &   0.13       &  16.85  &  4.88      &  2.40    &  10.53   &  0.94\\
\hspace*{3mm} VI        &  0.19   &       3.14 &   0.03  &    11.07     &  4.13     &   2.03     &   9.21      &  0.91\\
\hspace*{3mm} GVCL  &  \textbf{0.19}   & \textbf{2.98}         &   \textbf{0.02}       &  \textbf{9.51}   &  \textbf{3.99}       &     \textbf{1.94}    &  \textbf{8.96}           &  0.94\\ 
\hspace*{3mm} GVCL-Det  &  0.15  &   3.60       &  0.18        & 11.85   & \textit{4.55}        &    2.26    &  \textit{9.93}    &  0.91 \\ \\
50\% Data & & & & & & & & \\
\hspace*{3mm} Base       &  0.12   &  3.72       &  0.07       &  20.88   &  5.47      &  2.60  & 12.03   & 0.87\\
\hspace*{3mm} Transfer &  0.13 &  \textit{3.55} & 0.07 & 15.77   &   5.24  &  \textit{2.33}& 11.14   &  \textbf{\textit{0.96}} \\
\hspace*{3mm} Loss        &  0.11  &  3.90     &  \textit{ 0.05}      &  24.83  &  5.84  &  2.68 &  12.61   &  0.91\\\
\hspace*{3mm} VI        &  0.16  &       3.53  &   0.04 &    16.49   &  4.61      &   2.21  &   10.24      &  0.89\\
\hspace*{3mm} GVCL  &  \textbf{0.18}  &   \textbf{3.20}   &  \textbf{0.02}  &   \textbf{12.67}  &   \textbf{4.30}     & \textbf{1.63}    & \textbf{9.57}    & 0.92 \\ 
\hspace*{3mm} GVCL-Det  & \textit{0.14} &  3.75    &   0.13    &   \textit{15.67}    &  \textit{4.76}     &    2.41   & \textit{10.41}  & 0.91\\ \\
10\% Data & & & & & & & & \\
\hspace*{3mm} Base       & 0.09 & 4.46        &  \textit{0.06}        & 41.37   & 6.37     & 3.06   & 13.81    &   0.85 \\
\hspace*{3mm} Transfer & 0.08   & 4.36       &  0.12       &  32.84   &  6.41  &  3.14   &  13.41 &  \textit{\textbf{0.98}} \\
\hspace*{3mm} Loss        &  0.08  &  4.48      &  0.07      &  39.51 &  6.77   &  3.24   &  14.25 &  0.84 \\
\hspace*{3mm} VI       &  0.10 & 4.53 & 0.07   & 35.61   &    6.03  & 2.80 & 12.77    &  0.82\\
\hspace*{3mm} GVCL  &  \textbf{0.13} &  \textbf{3.86}    & \textbf{0.03}    & 27.73   &  \textbf{5.51}    &   \textbf{2.55}   &  \textbf{11.86}    &  0.91 \\
\hspace*{3mm} GVCL-Det  &  \textit{0.11} &  \textit{4.16}    &    0.13& \textit{\textbf{25.66}}   &    \textit{5.77}  &     \textit{2.75} & \textit{12.08}     &0.93\\
\end{tabular}
\end{center}
\end{table*}

\subsection{Results}

In what follows, the arithmetic mean of three independent runs for each experiment are reported. We also recreate the results from \citet{lit:covernet}, with our own hyperparameters, which partially improved upon the results reported in the original paper.

\subsection{Effect of Constrained Training Data}
\label{sec:results_data}

\begin{table*}[t]
\footnotesize
\caption{Results for different coverage bounds with 100\% of the  training data (\textbf{bold} as unique best result and \textit{italic} as unique best deterministic per set)} \label{tab:results_full}
\begin{center}
\begin{tabular}{l*{8}{R}}
\textbf{Experiment} & \textbf{ACC} & \textbf{NLL} & \textbf{ECE} & \textbf{RNK} & \ade[b]{1} & \ade[b]{5}& \fde[b]{1} & \textbf{DAC} \\
\hline \\
$\epsilon$=2 & & & & & & & & \\
\hspace*{3mm} Base       &  0.05   & 5.23         &  \textit{0.05}       & 106.04    &  5.21       &     2.69    & 11.67    & 0.89\\
\hspace*{3mm} Transfer &  0.04  &  \textit{5.19} &  0.06   & \textit{82.77}    & 5.38        &    2.54     &  11.50          & 0.97 \\
\hspace*{3mm} Loss        & 0.04   &    5.68      &    0.13     &  124.9   &  5.09       &    \textit{2.44}     &  10.76           &  0.97 \\
\hspace*{3mm} VI        &  0.06   &   5.13       &  \textbf{0.02}        & 98.43    &  4.91       &   2.30     &  10.87          & 0.89\\
\hspace*{3mm} GVCL  &   \textbf{0.06}  &  \textbf{4.94}        &   0.05       &   \textbf{74.12}  &    \textbf{4.60}     &  \textbf{2.26}       &  \textbf{10.24}             & 0.93\\ 
\hspace*{3mm} GVCL-Det  &   0.05  &7.53 &      0.36    &   123.93  &   \textit{4.87}      & 2.57        & \textit{10.43}         & 0.93 \\ \\
$\epsilon$=4 & & & & & & & & \\
\hspace*{3mm} Base       &  0.15   &  3.40       &  \textit{0.05}       &  15.07   &  4.77      &  2.26      & 10.61              & 0.88\\
\hspace*{3mm} Transfer &  0.15   &  \textit{3.24} & 0.06 & \textit{11.56}    &   4.70  &  \textit{2.12} & 10.15    &  \textbf{\textit{0.96}} \\
\hspace*{3mm} Loss        &  0.13   &  3.58       &   0.13       &  16.85  &  4.88      &  2.40       &  10.53         & 0.94\\
\hspace*{3mm} VI        &  0.19  &       3.14  &   0.03  &    11.07     &  4.13       &   2.03       &   9.21          & 0.91\\
\hspace*{3mm} GVCL  &  \textbf{0.19}   & \textbf{2.98}         &   \textbf{0.02}       &  \textbf{9.51}   &  \textbf{3.99}       & \textbf{1.40}     &  \textbf{8.96}            & 0.94\\ 
\hspace*{3mm} GVCL-Det  &  0.15   &   3.60       &  0.18        & 11.85   & \textit{4.55}        &    2.26     & \textit{9.93}             & 0.91 \\ \\
$\epsilon$=8 & & & & & & & & \\
\hspace*{3mm} Base       & 0.35 & 2.00         &  \textit{0.06}        & 3.45    & 4.95       & 2.43        & 10.86              & 0.85 \\
\hspace*{3mm} Transfer & 0.33    & \textit{1.95}        &  0.09        &  3.20   &  4.96       &  2.34       & 10.56            &  0.89 \\
\hspace*{3mm} Loss        &  0.35   &  1.98        &  0.08       &  3.43   &  4.91       &  2.40       &  10.62           & 0.88 \\
\hspace*{3mm} VI        & 0.41   &   1.84       &  \textbf{0.03}        &  2.94   &    4.33    &  2.31       & 9.44             &0.87 \\
\hspace*{3mm} GVCL  & \textbf{0.42}    & \textbf{1.73}         &  0.05        & \textbf{2.78}     &  \textbf{4.19}       &  \textbf{2.27}       & \textbf{9.16}            & \textbf{0.90} \\
\hspace*{3mm} GVCL-Det  & \textit{0.37}    &  2.10        &   0.19       &  \textit{3.08}   &  \textit{4.68}       &  \textit{2.33}       &  \textit{9.78}            & 0.89
\end{tabular}
\end{center}
\end{table*}
Tab.~\ref{tab:results_partial} evaluates the performance of our algorithm in comparison with the baseline models and focus on the effect of different amounts of available observations. 

We make the following observations. Our GVCL-CoverNet model outperforms all baselines in all metrics except the DAC, independent of the number of available observations. 
Most notably, the prediction performance of GVCL-CoverNet as measured by the ACC, \ade[n]{1} and \fde[n]{1} is substantially better compared to the other informed baselines (Transfer- and Loss-CoverNet). We also observe that GVCL-CoverNet shows a better calibration (NLL and ECE) and that the NLL and RNK gains significantly over the baselines increase with lesser amounts of data. This shows that our method is especially useful in case only limited observations are available.

In case of the DAC, our model is outperformed by the Transfer-CoverNet baseline. However, we hypothesize that Transfer-CoverNet may overly bias the DAC as it performs worse in all other metrics. While we can safely assume that the ground truth trajectory lies in the drivable area, we cannot assume the same for the respective positive label in trajectory set, due to the clustering (c.f.~\ref{sec:covernet}) Thus, our GVCL-CoverNet model may correctly trade-off the DAC for general performance gains, especially as measured by the ACC. We also note, that the Loss-CoverNet baseline is not able to consistently outperform Base-CoverNet even though it improves on the DAC as well.

A caveat is that our GVCL-CoverNet model employs double the parameters due to the variational formulation. However, we observe that our GVCL-Det-CoverNet still achieves the best ACC, \ade[n]{1}, \fde[n]{1} of all non-variational models. 
The RNK and \ade[n]{5} show worse performance than Transfer-CoverNet, possible because substantial calibration is lost when disregarding the already learned variance information. However, with decreasing amounts of available observations, the GVCL-Det-CoverNet mode becomes more competitive. In fact, we can obtain the predictive performance (in terms of \ade[n]{1} and \fde[n]{1}) of the Base model, with just $50\%$ of the training data.
This further substantiates the usefulness of our method under limited amounts of observations. 

It should be noted, the performance gains can only be partially attributed to the probabilistic formulation. The non-informed VI-CoverNet model already improves on the Base-CoverNet model across all metrics, but GVCL-CoverNet again outperforms the VI-CoverNet model across all metrics. Most notably, the calibration as measured by the NLL and ECE is increasingly improved the less observations are available. We see according improvements in the calibration-senstive RNK and \ade[n]{5} metrics and a higher DAC, especially when only 10\% of observations are available. 

\subsection{Effect of Coverage Bound}

\label{sec:results_epsilon}

Tab.~\ref{tab:results_full} investigates the effect of the coverage bounds $\epsilon \in \{2,4,8\}$ on the prediction performance, using $100\%$ of the training data. This allows a closer comparison to the basic CoverNet approach, that was also evaluated with multiple $\epsilon$ values.

Similar as before, we first observe that our GVCL-CoverNet model still outperforms all informed baselines for each coverage bound across all metrics except the DAC. 
Regarding the ECE, the VI-CoverNet model achieves a better performance for coverage bounds $\epsilon \in \{2,8\}$. However, the ECE is here not consistent with the NLL, which makes it difficult to draw conclusions about the calibration of the two models. As in Sec.~\ref{sec:results_data}, we also observe that our GVCL-Det-CoverNet model achieves substantial performance gains with the best ACC, \ade[n]{1}, \fde[n]{1} of all non-variational models in all sets of experiments. However, we see here that the calibration issues of the GVCL-Det-CoverNet model persist, with the worst NLL and ECE in all sets of experiments. We again conclude, that it is not optimal to simply discard the variance information.

Second, we observe that all models achieve their respective best prediction performance (as measured by \ade[n]{1}, \ade[n]{5} and \fde[n]{1}) with a coverage bound $\epsilon=4$. We assume that a low trajectory set resolution with a coverage bound $\epsilon=8$ is too coarse to sufficiently cover all possible trajectories, leading to higher deviation from the actual ground truth, while a high resolution with coverage bound $\epsilon=2$ poses a too difficult prediction challenge. 


\section{Conclusion}
\label{sec:conclusion}

We propose a probabilistic continual learning approach to integrate prior knowledge as soft constraint into deep learning models, without assuming any specific architecture. We introduce an exemplary GVCL-CoverNet application, that integrates prior drivable area knowledge, using already available map data, in the multi-modal trajectory prediction for autonomous driving. Our implementation substantially outperforms the original CoverNet, even when trained with only half the observations. Furthermore, it outperforms existing informed CoverNet baselines, with increasing benefits when only a limited amount of observations is available, even when accounting for advantages of the probabilistic approach. In result, we clearly demonstrate the benefits of informing the trajectory prediction through prior knowledge and that a continual variant is to be preferred over loss- or transfer-based methods.

Future research might explore other probabilistic continual learning methods, e.g. with subspace approximations, or the integration of multiple prior knowledge sources, e.g.~lane markings and traffic signals in the trajectory prediction. Furthermore, the optimal method for deriving a deterministic variant of an probabilistic model should be analysed. Another direction are more flexible forms of variational densities, such as those allowing for posterior dependencies in the variational covariance.




\subsubsection*{Acknowledgements}
The research leading to these results is funded by the German Federal Ministry for Economic Affairs and Climate Action within the project “KI Wissen – Entwicklung von Methoden für die Einbindung von Wissen in maschinelles Lernen". The authors would like to thank the consortium for the successful cooperation.

\bibliography{references}

\clearpage
\appendix

\thispagestyle{empty}

\makesupplementtitle

\section{Hyperparameter Tuning}

As reported in the main paper, we individually tune the learning rates and batch sizes for each model as well as the $\lambda_\text{\scriptsize{multi}}$ hyperparameter for the Loss-CoverNet baseline. The tuning is carried out with 100\% of the observational training data with coverage bounds $\epsilon \in \{2,8\}$ respectively. The values are selected to minimize the validation negative log likelihood (NLL) using the heteroscedastic and evolutionary Bayesian optimizer (HEBO). Here, we summarize the respective tuning ranges and resulting values for the learning rate and batch size in Tab.~\ref{tab:hypparams_full}. The hyperparameter $\lambda_{\scriptsize{\text{multi}}}$ is tuned in the range $0.01 \dots 1.0$ with $0.01$ selected for both coverage bounds. Note, that we adopt the results from tuning with $\epsilon=8$ for the experiments with $\epsilon=4$. 

\begin{table}[h!]
\footnotesize
\caption{Selected values and tuning ranges for batch size and learning rate.} \label{tab:hypparams_full}
\begin{center}
\begin{tabular}{lrrrr}
\textbf{Method} & \textbf{Value Batch Size}& \textbf{Range Batch Size} & \textbf{Value Learning Rate}&\textbf{Range Learning Rate}\\
\hline \\
$\epsilon$=2  & & &   & \\  
\hspace*{3mm} Base       &  5  & 4 $\dots$ 32  & 0.00083    & 0.00001 $\dots$ 0.001 \\
\hspace*{3mm} Transfer &  5   & 4 $\dots$ 32 &  0.00083  & 0.00001 $\dots$ 0.001\\
\hspace*{3mm} Loss        & 5 &  4 $\dots$ 32 & 0.00083 & 0.00001 $\dots$ 0.001 \\
\hspace*{3mm} VI        &   12 & 4 $\dots$16 & 0.003 &  0.001 $\dots$ 0.05  \\
\hspace*{3mm} GVCL  &   12 & 4 $\dots$16 & 0.003 & 0.001 $\dots$ 0.05 \\ \\
$\epsilon$=8  & & &   & \\                
\hspace*{3mm} Base     &   16 & 4 $\dots$ 16 & 0.0008& 0.00001 $\dots$ 0.001\\
\hspace*{3mm} Transfer & 16 &  4 $\dots$ 16& 0.0002&0.00001 $\dots$ 0.001\\
\hspace*{3mm} Loss        &  16 &   4 $\dots$ 16 & 0.0001 & 0.00001 $\dots$ 0.001 \\
\hspace*{3mm} VI        &   12 &   4 $\dots$ 16 & 0.003  &   0.001 $\dots$ 0.05 \\
\hspace*{3mm} GVCL  &   12 & 4 $\dots$ 16& 0.003  &0.001 $\dots$ 0.05
\end{tabular}
\end{center}
\end{table}

\section{Additional Metrics}

In addition to the metrics in our main paper results, we evaluated the \ade[n]{10}, \ade[n]{15} and \hrate[n]{5}{2} metrics (Huang et al., 2022). The \ade[n]{10} and \ade[n]{15} evaluate the quality of the 10 and 15 most likely trajectories in the prediction respectively. The \hrate[n]{5}{2} measures if the 5 most likely trajectories are, roughly said, close enough to the ground truth, given a 2 meter threshold. Similar to the \ade[n]{5} metric, these additional metrics indirectly depend on well calibrated, probabilistic predictions.

We report here a complete summary of the results including these metrics in Tab.~\ref{tab:results_partial_supplemental} regarding the experiments with constrainted observational training datasets and in Tab.~\ref{tab:results_full_supplemental} regarding the experiments with different coverage bounds $\epsilon \in \{2, 4, 8\}$. Furthermore, we include here in both tables the standard deviation of the three independent experiment runs. We mainly omitted this information in our main paper to stay concise. In particular, we observe that the \ade[n]{10} and \ade[n]{15} behave similar to the \ade[n]{5} metric we discussed in our main paper. We also see that the \hrate[n]{5}{2} metric behaves similar to the discussed \ade[n]{1} and \fde[n]{1} metrics. Regarding the standard deviation, we observe that the Base-CoverNet model exhibits the most variation. This shows that the prediction quality of the basic CoverNet can differ substantially.

\begin{table}[t]
\tiny
\caption{Results for constrained observational training datasets with $\epsilon=4$ (\textbf{bold} as unique best result and \textit{italic} as unique best deterministic per set)} \label{tab:results_partial_supplemental}
\begin{center}
\begin{tabular}{l*{12}{R}}
\textbf{Experiment} & \textbf{Top1 ACC} & \textbf{NLL} & \textbf{ECE} & \textbf{RNK} & \ade[b]{1} & \ade[b]{5}& \ade[b]{10} & \ade[b]{15} & \fde[b]{1} & \hrate[b]{5}{2} & \textbf{DAC} \\
\hline \\
100\%   & & & & & & & & & & \\
\hspace*{3mm} Base       &  0.15\tpm 0.02   &  3.40\tpm0.1        &  0.05\tpm0.02        &  15.07\tpm1.98   &  4.77\tpm0.49       &  2.26\tpm0.10       &   1.74\tpm0.04       &  1.56\tpm0.03        & 10,61\tpm1.16    &   0.19\tpm0.01          & 0.88\tpm 0.01\\
\hspace*{3mm}Transfer &  0.15\tpm0.00   &  \textit{3.24\tpm0.01} & 0.06\tpm0.00 & \textit{11.56\tpm0.09}    &   4.70\tpm0.04   &  \textit{2.12\tpm0.01} &  \textit{1.61\tpm0.01}    & \textit{1.44\tpm0.01}   & 10.15\tpm0.06    &  0.19\tpm0.00  &  \textbf{\textit{0.96\tpm 0.01}} \\
\hspace*{3mm} Loss        &  0.13\tpm0.01   &  3.58\tpm0.01        &   0.13\tpm0.02       &  16.85\tpm1.49   &  4.88\tpm0.13       &  2.40\tpm0.05       &    1.74\tpm0.04      &   1.55\tpm0.03       &  10.53\tpm0.27   &     0.18\tpm0.01        & 0.94\tpm0.01\\
\hspace*{3mm}VI        &  0.19\tpm0.00   &       3.14\tpm0.04  &   0.03\tpm0.00  &    11.07\tpm0.52     &  4.13\tpm0.13       &   2.03\tpm0.09       &    1.60\tpm0.06      &    1.45\tpm0.04 &   9.21\tpm0.35        & 0.21\tpm0.00  & 0.91\tpm0.01\\
\hspace*{3mm} GVCL  &  \textbf{0.19\tpm0.00}   & \textbf{2.98\tpm0.02}         &   \textbf{0.02\tpm0.00}       &  \textbf{9.51\tpm0.14}   &  \textbf{3.99\tpm0.05}       &     \textbf{1.94\tpm0.03}    &  \textbf{1.54\tpm0.02}        & \textbf{1.40\tpm0.01}     &  \textbf{8.96\tpm0.01}           & \textbf{0.22\tpm0.00} & 0.94\tpm0.01\\ 
\hspace*{3mm}GVCL Det  &  0.15\tpm0.01   &   3.60\tpm0.08       &  0.18\tpm0.02        & 11.85\tpm0.48    & \textit{4.55\tpm0.11}        &    2.26\tpm0.05     &  1.72\tpm0.02        &     1.52\tpm0.01     & \textit{9.93\tpm0.25}    &   0.19\tpm0.00          & 0.91\tpm0.02 \\ \\
50\%   & & & & & & & & & &  \\
\hspace*{3mm} Base       &  0.12\tpm 0.01   &  3.72\tpm0.09        &  0.07\tpm0.01        &  20.88\tpm1.01   &  5.47\tpm0.27       &  2.60\tpm0.12       &   1.92\tpm0.06       &  1.67\tpm0.04        & 12.03\tpm0.69    &   0.16\tpm0.01          & 0.87\tpm 0.01\\
\hspace*{3mm} Transfer &  0.13\tpm0.01   &  \textit{3.55\tpm0.02} & 0.07\tpm0.00 & 15.77\tpm0.18    &   5.24\tpm0.11   &  \textit{2.33\tpm0.03} &  \textit{1.75\tpm0.02}    & \textit{1.54\tpm0.01}   & 11.14\tpm0.33    &  0.17\tpm0.01  &  \textbf{\textit{0.96\tpm 0.01}} \\
\hspace*{3mm} Loss        &  0.11\tpm0.00   &  3.90\tpm0.02        &   \textit{0.05\tpm0.02}       &  24.83\tpm0.32   &  5.84\tpm0.15       &  2.68\tpm0.04       &    1.98\tpm0.03      &   1.72\tpm0.01       &  12.61\tpm0.25   &     0.15\tpm0.00        & 0.91\tpm0.01\\
\hspace*{3mm} VI        &  0.16\tpm0.00   &       3.53\tpm0.02  &   0.04\tpm0.01  &    16.49\tpm0.15     &  4.61\tpm0.01       &   2.21\tpm0.02       &    1.73\tpm0.01      &    1.56\tpm0.01 &   10.24\tpm0.03        & 0.19\tpm0.00  & 0.89\tpm0.01\\
\hspace*{3mm} GVCL  & \textbf{0.18\tpm0.00}  & \textbf{3.20\tpm0.02}     & \textbf{0.02\tpm0.01}   & \textbf{12.67\tpm0.45}   & \textbf{4.30\tpm0.03}     & \textbf{2.08\tpm0.01}     &  \textbf{1.63\tpm0.01}      &  \textbf{1.48\tpm0.01}   &   \textbf{9.57\tpm0.01}   & \textbf{0.20\tpm0.00} & 0.92\tpm0.00 \\ 
\hspace*{3mm} GVCL Det  &  \textit{0.14\tpm0.01} &  3.75\tpm0.09    &   0.13\tpm0.01    &   \textit{15.67\tpm0.89} &  \textit{4.76\tpm0.16}     &   2.41\tpm0.09  &  1.84\tpm0.06     &  1.62\tpm0.05     & \textit{10.41\tpm0.42}  &  \textit{0.18\tpm0.01}     & 0.91\tpm0.00  \\ \\
10\%  & & & & & & & & & & \\
\hspace*{3mm} Base       & 0.09\tpm0.01 & 4.46\tpm0.05         &  \textit{0.06\tpm0.01}        & 41.37\tpm1.92    & 6.37\tpm0.17        & 3.06\tpm0.08        & 2.26\tpm0.06         &  1.93\tpm0.05        & 13.81\tpm0.33    &   0.13\tpm0.01          & 0.85\tpm0.00 \\
\hspace*{3mm} Transfer & 0.08\tpm0.01    & 4.36\tpm0.08        &  0.12\tpm0.01        &  32.84\tpm1.44   &  6.41\tpm0.14       &  3.14\tpm0.10       &  2.35\tpm0.19        &  1.90\tpm0.04        & 13.41\tpm0.19    &   0.13\tpm0.00          &  \textit{\textbf{0.98\tpm0.00}} \\
\hspace*{3mm} Loss        &  0.08\tpm0.00   &  4.48\tpm0.07        &  0.07\tpm0.01        &  39.51\tpm1.08   &  6.77\tpm0.46       &  3.24\tpm0.14       & 2.38\tpm0.10          &  2.01\tpm0.06        &  14.25\tpm0.19   &   0.11\tpm0.01          & 0.84\tpm0.03 \\
\hspace*{3mm} VI       &  0.10\tpm0.00 & 4.53\tpm0.06  & 0.07\tpm0.01    & 35.61\tpm1.58   &    6.03\tpm0.15  & 2.80\tpm0.08     &  2.06\tpm0.04      & 1.79\tpm0.02    & 12.77\tpm0.22     & 0.14\tpm0.00 & 0.82\tpm0.02\\
\hspace*{3mm} GVCL  &  \textbf{0.13\tpm0.01} &  \textbf{3.86\tpm0.05}    & \textbf{0.03\tpm0.00}    & 27.73\tpm1.40   &  \textbf{5.51\tpm0.12}    &   \textbf{2.55\tpm0.06}   &   \textbf{1.94\tpm0.02}     &  \textbf{1.69\tpm0.01}   &  \textbf{11.86\tpm0.29}    & \textbf{0.17\tpm0.00} & 0.91\tpm0.01 \\
\hspace*{3mm} GVCL Det  &  \textit{0.11\tpm0.01} &  \textit{4.16\tpm0.15}    &    0.13\tpm 0.03& \textit{\textbf{25.66\tpm2.67}}   &    \textit{5.77\tpm0.07}  &     \textit{2.75\tpm0.05} & \textit{2.06\tpm0.08}       &     \textit{1.79\tpm0.06} & \textit{12.08\tpm0.09}     & \textit{0.16\tpm0.00} & 0.93\tpm0.02\\
\end{tabular}
\end{center}
\end{table}

\begin{table}[b]
\tiny
\caption{Results for different coverage bounds with 100\% of the observational training dataset (\textbf{bold} as unique best result and \textit{italic} as unique best deterministic per set)} \label{tab:results_full_supplemental}
\begin{center}
\begin{tabular}{l*{12}{R}}
\textbf{Experiment} & \textbf{Top1 ACC} & \textbf{NLL} & \textbf{ECE} & \textbf{RNK} & \ade[b]{1} & \ade[b]{5}& \ade[b]{10} & \ade[b]{15} & \fde[b]{1}  & \hrate[b]{5}{2} & \textbf{DAC} \\
\hline \\
$\epsilon$=2  & & & & & & & & & & \\
\hspace*{3mm} Base       &  0.05\tpm0,00   & 5.23\tpm0.03         &  0.05\tpm0.02        & 106.04\tpm5.75    &  5.21\tpm0.21       &     2.69\tpm0.21    &  1.89\tpm0.03         &  1.62\tpm0.01        & 11.67\tpm0.44     &   0.26\tpm0.00          & 0,89\tpm0,00\\
\hspace*{3mm} Transfer &  0.04\tpm0.00   &  \textit{5.19\tpm0.02} &  \textit{0.06\tpm0.01}   & \textit{82.77\tpm2.06}    & 5.38\tpm0.13        &    2.54\tpm0.04     &  1.83\tpm0.03 &   \textit{1.52\tpm0.02}       &  11.50\tpm0.28   &     0.25\tpm0.00        & 0.97\tpm0.00 \\
\hspace*{3mm}  Loss        & 0.04\tpm0.00    &    5.68\tpm0.07      &    0.13\tpm0.01      &  124.9\tpm1.16   &  5.09\tpm0.04       &    \textit{2.44\tpm0.04}     &    \textit{1.81\tpm0.03}      &   1.55\tpm0.02       &  10.76\tpm0.09   &    0.23\tpm0.00         &  0.97\tpm 0.01 \\
\hspace*{3mm}  VI        &  0.06\tpm0.00   &   5.13\tpm0.01       &  \textbf{0.02\tpm0.00}        & 98.43\tpm1.72    &  4.91\tpm0.1       &   2.30\tpm0.01      &  1.71\tpm0.02        & 1.46\tpm0.02         &  10.87\tpm0.24   &     \textbf{0.31\tpm0.01}        & 0.89\tpm0.00\\
\hspace*{3mm}  GVCL  &   \textbf{0.06\tpm0.00}  &  \textbf{4.94\tpm0.01}        &   0.05\tpm0.00       &   \textbf{74.12\tpm0.95}  &    \textbf{4.60\tpm0.09}     &  \textbf{2.26\tpm0.01}       &     \textbf{1.68\tpm0.01}     &  \textbf{1.42\tpm0.01}        & \textbf{10.24\tpm0.23}    &   0.30\tpm0.01          & 0.93\tpm0.01\\ 
\hspace*{3mm}  GVCL Det  &   0.05\tpm0.00  &7.53\tpm0.59 &      0.36\tpm0.08    &   123.93\tpm11.52  &   \textit{4.87\tpm0.09}      & 2.57\tpm0.07        &  1.92\tpm0.07        &  1.62\tpm0.06        & \textit{10.43\tpm0.12}    &       0.25\tpm0.01      & 0.93\tpm0.00 \\ \\
$\epsilon$=4  & & & & & & & & & & \\
\hspace*{3mm} Base       &  0.15\tpm 0.02   &  3.40\tpm0.1        &  0.05\tpm0.02        &  15.07\tpm1.98   &  4.77\tpm0.49       &  2.26\tpm0.10       &   1.74\tpm0.04       &  1.56\tpm0.03        & 10,61\tpm1.16    &   0.19\tpm0.01          & 0.88\tpm 0.01\\
\hspace*{3mm}  Transfer &  0.15\tpm0.00   &  \textit{3.24\tpm0.01} & 0.06\tpm0.00 & \textit{11.56\tpm0.09}    &   4.70\tpm0.04   &  \textit{2.12\tpm0.01} &  \textit{1.61\tpm0.01}    & \textit{1.44\tpm0.01}   & 10.15\tpm0.06    &  0.19\tpm0.00  &  \textit{0.96\tpm 0.01} \\
\hspace*{3mm} Loss        &  0.13\tpm0.01   &  3.58\tpm0.01        &   0.13\tpm0.02       &  16.85\tpm1.49   &  4.88\tpm0.13       &  2.40\tpm0.05       &    1.74\tpm0.04      &   1.55\tpm0.03       &  10.53\tpm0.27   &     0.18\tpm0.01        & 0.94\tpm0.01\\
\hspace*{3mm} VI        &  0.19\tpm0.00   &       3.14\tpm0.04  &   0.03\tpm0.00  &    11.07\tpm0.52     &  4.13\tpm0.13       &   2.03\tpm0.09       &    1.60\tpm0.06      &    1.45\tpm0.04 &   9.21\tpm0.35        & 0.21\tpm0.00  & 0.91\tpm0.01\\
\hspace*{3mm} GVCL  &  \textbf{0.19\tpm0.00}   & \textbf{2.98\tpm0.02}         &   \textbf{0.02\tpm0.00}       &  \textbf{9.51\tpm0.14}   &  \textbf{3.99\tpm0.05}       &     \textbf{1.94\tpm0.03}    &  \textbf{1.54\tpm0.02}        & \textbf{1.40\tpm0.01}     &  \textbf{8.96\tpm0.1}           & \textbf{0.22\tpm0.00} & 0.94\tpm0.01\\ 
\hspace*{3mm} GVCL Det  &  0.15\tpm0.01   &   3.60\tpm0.08       &  0.18\tpm0.02        & 11.85\tpm0.48    & \textit{4.55\tpm0.11}        &    2.26\tpm0.05     &  1.72\tpm0.02        &     1.52\tpm0.01     & \textit{9.93\tpm0.25}    &   0.19\tpm0.00          & 0.91\tpm0.02 \\ \\
$\epsilon$=8   & & & & & & & & & & \\
\hspace*{3mm} Base       & 0.35\tpm0.01 & 2.00\tpm0.01         &  \textit{0.06\tpm0.01}        & 3.45\tpm0.04    & 4.95\tpm0.09        & 2.43\tpm0.04        & 2.17\tpm0.01         &  2.11\tpm0.01        & 10.86\tpm0.24    &   0.08\tpm0.00          & 0.85\tpm0.01 \\
\hspace*{3mm}  Transfer & 0.33\tpm0.01    & \textit{1.95\tpm0.01}        &  0.09\tpm0.03        &  3.20\tpm0.02   &  4.96\tpm0.1       &  2.34\tpm0.02       &  2.13\tpm0.00        &  2.09\tpm0.00        & 10.56\tpm0.13    &   0.08\tpm0.00          &  0.89\tpm0.02 \\
\hspace*{3mm}  Loss        &  0.35\tpm0.01   &  1.98\tpm0.01        &  0.08\tpm0.02        &  3.43\tpm0.06   &  4.91\tpm0.14       &  2.40\tpm0.05       & 2.16\tpm0.01          &  2.11\tpm0.00        &  10.62\tpm0.31   &   0.08\tpm0.00          & 0.88\tpm0.02 \\
\hspace*{3mm}  VI        & 0.41\tpm0.00    &   1.84\tpm0.04       &  \textbf{0.03\tpm0.01}        &  2.94\tpm0.07   &    4.33\tpm0.09     &  2.31\tpm0.01       &  2.13\tpm0.00        &    2.10\tpm0.00      & 9.44\tpm0.22    &   0.08\tpm0.00          &0.87\tpm0.00 \\
\hspace*{3mm}  GVCL  & \textbf{0.42\tpm0.00}    & \textbf{1.73\tpm0.01}         &  0.05\tpm0.01        & \textbf{2.78\tpm0.01}     &  \textbf{4.19\tpm0.03}       &  \textbf{2.27\tpm0.01}       &  \textbf{2.11\tpm0.01}        &  \textbf{2.09\tpm0.00}        & \textbf{9.16\tpm0.04}    &    0.08\tpm0.00         & \textbf{0.90\tpm0.01} \\
\hspace*{3mm}  GVCL Det  & \textit{0.37\tpm0.01}    &  2.10\tpm0.01        &   0.19\tpm0.02       &  \textit{3.08\tpm0.04}   &  \textit{4.68\tpm0.01}       &  \textit{2.33\tpm0.01}       &  \textit{2.12\tpm0.00}        &  \textit{2.09\tpm0.00}        &  \textit{9.78\tpm0.10}   &    0.08\tpm0.00         & \textit{0.89\tpm0.01}
\end{tabular}
\end{center}
\end{table}

\end{document}